\documentclass[runningheads]{llncs}

\usepackage{graphicx}
\usepackage{times} 
\usepackage{helvet}
\usepackage{courier} 
\usepackage[hyphens]{url} 
\usepackage{graphicx}
\usepackage{verbatim}
\usepackage{caption}
\usepackage{multirow}
\usepackage{xcolor,colortbl}

\begin{document}
\title{Clinical Dialogue Transcription Error Correction using Seq2Seq Models}

\author{Gayani Nanayakkara \and
Nirmalie Wiratunga \and
David Corsar \and Kyle Martin \and Anjana Wijekoon}
\authorrunning{G. Nanayakkara et al.}
%
\institute{Robert Gordon University, Aberdeen, Scotland \\
\email{\{g.nanayakkara, n.wiratunga, d.corsar1, k.martin3, a.wijekoon1\}@rgu.ac.uk}}

\definecolor{Gray}{gray}{0.6}
\definecolor{gainsboro}{rgb}{0.86, 0.86, 0.86}
\definecolor{black}{rgb}{0.0, 0.0, 0.0}

\maketitle
\begin{abstract}
Good communication is critical to good healthcare. 
Clinical dialogue is a conversation between health practitioners and their patients, with the explicit goal of obtaining and sharing medical information. 
This information contributes to medical decision-making regarding the patient and plays a crucial role in their healthcare journey. 
The reliance on note taking and manual scribing processes are extremely inefficient and leads to manual transcription errors when digitizing notes. 
Automatic Speech Recognition (ASR) plays a significant role in speech-to-text applications, and can be directly used as a text generator in conversational applications. 
However, recording clinical dialogue presents a number of general and domain-specific challenges.
In this paper, we present a seq2seq learning approach for ASR transcription error correction of clinical dialogues. 
We introduce a new Gastrointestinal Clinical Dialogue (GCD) Dataset which was gathered by healthcare professionals from a NHS Inflammatory Bowel Disease clinic and use this in a comparative study with four commercial ASR systems. 
Using self-supervision strategies, we fine-tune a seq2seq model on a mask-filling task using a domain-specific PubMed dataset which we have shared publicly for future research. 
The BART model fine-tuned for mask-filling was able to correct transcription errors and achieve lower word error rates for three out of four commercial ASR outputs.

\end{abstract}

\section{Introduction}
Traditional approaches to record keeping in a health service setting have relied on pen to paper for all clinical professionals who ask the same questions of the same patient. 
Drawbacks to this approach include the time burden of record keeping of clinical communications, the potential for error and most importantly means the patient more often than not is required to repeat and share the same detail asked in a different way.  
The absence of accurate clinical dialogue capture is a contributory factor to poor communication in medical practice~\cite{McDonald_2016}. 
To a patient this promotes mistrust and a feeling of fragmented care in a system that seems not to be linked up.

Clinical documentation is time consuming and is associated with clinician burnout, increased cognitive load, information loss, and distractions~\cite{Quiroz2019}. 
One of the most promising avenues of automating clinical documentation with digital scribes is to use an Automatic Speech Recognition (ASR)~\cite{Radford2019LanguageMA} system; whereby in a process called digital transcription, audio data received as input is converted to textual data as output. Recent advances in Natural Language Processing~(NLP) and adoption of cloud-based technologies have created a significant market for ASR systems.
Due to the critical nature of the domain, ASR for clinical applications are expected to demonstrate high levels of performance. 
However variations in language, speech and environmental contexts 
have made it hard to achieve an acceptable levels of transcription accuracy~\cite{asr_review}. 
Thus, it is important to examine strategies to mitigate or reduce the likelihood of error in a transcription. 

There are two approaches to correcting ASR errors: redesign and retrain the core ASR architecture; or alternatively perform a post-ASR error correction on the transcribed ASR output. In this paper we focus on the second approach and use a seq2seq fine-tuned neural model to map an ASR transcribed piece of text to its error corrected form.  
We select T5~\cite{T5} and BART~\cite{lewis2019bart} as our seq2seq models due to their dominant performance across domains. 
A self-supervised training strategy with fine-tuning tasks is used with a novel domain-specific dataset scraped from PubMed~\footnote{https://www.ncbi.nlm.nih.gov/pubmed/} abstracts. 
We identified a lack of specific clinical dialogue datasets in related literature and 
address this deficit by introducing a novel clinical dialogue dataset which is used to test the effectiveness of our error correction models. 
Results from a comparative study of seq2seq models show that our proposed approach can reduce transcription errors that are introduced by several commercial ASR systems.
Accordingly, our contributions are:
\begin{itemize}
 \item We demonstrate clinical dialogue error correction using the Gastrointestinal Clinical Dialogue~(GCD) Dataset which we gathered in partnership with National Health Service~(NHS) Scotland;
 \item A self-supervision methodology to fine-tune language models for clinical dialogue error corrections using a novel PubMed dataset; and
 \item A comparative evaluation of fine-tuned language models for clinical dialogue transcription error correction.
\end{itemize}

The rest of the paper is organised as follows. Section~\ref{sec:related} presents related literature in ASR error correction methods. The Gastrointestinal Clinical Dialogue~(GCD) dataset is presented in Section~\ref{sec:nhsdata} followed by Section~\ref{sec:methods} which presents the language models considered for error correction and our approach to fine-tuning language models using the self-supervised PubMed datasets. 
Section~\ref{sec:eval} presents the comparative evaluation of language models and fine-tuned models for error correction using the GCD dataset. 
In Section~\ref{sec:discuss} we further investigate the performance improvements we observed in the previous section to draw insights for future work. Finally we present our conclusions in Section~\ref{sec:conc}.

\section{Related Work} \label{sec:related}

ASR techniques are used to capture real-time speech in audio format and convert them into textual outputs. 
In clinical settings, ASR can be used as the initial step to gather the conversational data and to produce meaningful insights from the generated ASR transcriptions. 
However, ASR performance is mainly dependent on three factors: speaker variabilities~(changes in voice due to ageing, illness, emotions and tiredness); spoken language variabilities~(variants in speech due to accents and dialects); and other mismatch factors~(communication channels and the devices)~\cite{asr_review}. 
Moreover, these factors affect the performance of the ASR systems, and will generate erroneous results, from which it is challenging to extract meaningful insights. The types of errors found in speech recognition are threefold: insertion, deletion, and substitution~\cite{asr_review}. Word Error Rate~(WER) is the common evaluation metric used to evaluate the performance of ASR outputs considering the three errors as mentioned above~\cite{asr_review,asr_benchmark}.

There are two alternative approaches for the ASR error correction: implement error correction algorithm within the ASR model; or as a post-processing step where the ASR outputs will be analysed for error correction. 
Hidden Markov Models~\cite{Humphries1996UsingAP,Kamper2011} and more recently deep neural architectures~\cite{Jain2018ImprovedAS} have been explored for ASR models that include error correction. 
The alternative (and increasingly more common) approach involving post-ASR error corrections have in the past adopted unsupervised approaches. Early methods include lexical co-occurrence analysis on large ASR transcription corpora~\cite{context} and using statistical error correction methods~\cite{cucu2013}. 
FastCorrect is a more recent transformer based architecture which integrates the edit distance metric within a deep neural architecture to guide error correction~\cite{leng2021fastcorrect}. 
Alternatively, transformer based architectures has been fine-tuned for error correction using part of the domain specific dataset~(a train set)~\cite{mani2020asr}. Increasingly, for post-ASR error correction there is potential to exploit recent advances in language modelling.
Accordingly, in this study, we will also focus on post-ASR error correction using a transformer based architectures. However, instead of implementing a customised architecture~\cite{leng2021fastcorrect} or fine-tuning with clinical dialogue data~(of which there is only a very limited amount of data)~\cite{mani2020asr}, we explore how to effectively fine-tune a pre-trained model using publicly available clinical domain data. 

\section{Clinical Dialogue Transcription} \label{sec:nhsdata}
Clinical dialogue is a conversation, typically between a clinician and a patient, with the explicit goal of obtaining medical information regarding the patient. This information contributes to medical decision-making regarding the patient and plays a crucial role in their healthcare journey. The clinician will ask questions about the patient's medical state, including: personal details, medical history, symptoms, prescribed medicines and clinical tests. Recording these conversations often relies on the clinician maintaining paper-based notes, which in turn requires effort and time when converting to patient Electronic Health Records (EHRs). Once created, EHRs are maintained by a centralised system and can be shared among all of the different medical specialists which may take part in a patient's care journey. To clarify, consider the following fictitious example. A patient creates an appointment with their local general practice due to complaints of a stomach ache. The general practice clinician then uses the information in the patient's EHR to check for allergies before prescribing new medication. Information regarding the prescription is then appended to the patient's EHR and is visible during a follow-on appointment with a gastrointestinal specialist and allows them to try a new treatment without repetition. Accurate and timely EHRs are therefore crucial to the success of patient treatment. Accordingly, there is a demand for algorithms that efficiently create accurate electronic health records from clinical dialogues. 

Manual note-taking is comparatively an inefficient process compared to digitisation of clinical dialogues. Clinicians lose valuable time on administrative tasks which could be put to better use elsewhere. Furthermore, there is the possibility of error or misunderstandings being introduced at the note-taking and digitization stages. 
Speech-to-text transformation presents an opportunity to create (or update) EHRs from clinical dialogues. 
However, speech-to-text conversion in a clinical setting presents a number of challenges, both general (i.e.  accurate transcription in the presence of background noise, different speaker accents and dialects, interruptions and repetitions) and domain-specific (i.e. recognising expert vocabulary, non-overlapping expertise between conversation participants). 
Once corrected, the goal is to extract clinical data and insights to formulate summaries of conversations that can then be captured as part of the an electronic patient health record. 

To understand the performance of speech-to-text transcription of clinical dialogues we require data specifically in the form of audio and its reference transcript. To the best of our knowledge, there is no available clinical conversational dataset in the English language. 
Here we describe an applied machine learning task using the Gastrointestinal Clinical Dialogue~(GCD) dataset which has been collected working in partnership with the National Health Service (NHS) Scotland. 
In this work we limit our scope to gastrointestinal disease related clinical conversations, which mainly took place in the Inflammatory Bowel Disease~(IBD) clinic. 

\subsection{Gastrointestinal Clinical Dialogue Dataset}
The clinical conversations in the GCD dataset were generated using role-playing conversations initiated in the NHS IBD Clinic. These conversations contain clinical dialogues that often take place between an IBD clinicians and a patient. The data collection included 7 participants with Scottish accents. The accent can be viewed as a form of noise in addition to common noise factors such as background noise, interruptions and repetitions. Overall, we collected 7 audio files each with $4\sim5$ minutes of conversation. Each audio file contains a mean number of 47 utterances where two persons engaged in a clinical conversation. A summary of audio data statistics are presented in Table~\ref{tbl:asrsummary}.
\begin{table}[t]
\centering
\renewcommand{\arraystretch}{1.1}
\caption{Summary of the GCD Dataset}
\label{tbl:asrsummary} 
\begin{tabular}{ll}
\hline 
\textbf{Feature}&\textbf{Value}\\
\hline
No. of audio files &\textbf{7}\\
Mean length of an audio file &\textbf{4 mins 49 secs}\\
Mean no. of utterances in a file &\textbf{47}\\
Mean no. of words in an utterance &\textbf{93}\\
\hline
\end{tabular}
\end{table}

The GCD dataset consist of 329~($\sim$47*7) data instances where each data instance has three components; audio file, reference transcript~(i.e. gold standard) and multiple ASR transcripts. 
The reference transcript is created by listening to the audio and manually transcribing it. 
To create the ASR transcriptions, we used four commercially available ASR systems. These include AWS Transcribe, Microsoft speech-to-text, IBM Watson and Google speech-to-text. These ASR services were selected from a large number of ASR services both commercial and open source based on their support for British English accent and popularity. 
Table~\ref{tbl:gcdexamples} presents some examples of reference transcriptions and their ASR Transcribe outputs~(AWS Transcribe in the example) from the GCD dataset.
\begin{table*}[t]
\centering
\renewcommand{\arraystretch}{1.3}
\caption{Examples from the GCD Dataset}
\label{tbl:gcdexamples} 
\begin{tabular}{p{6cm}p{6cm}}
\hline 
\textbf{Gold Reference}&\textbf{AWS Transcribe Output}\\
\hline
So do you have any ideas as to what might be the cause of your symptoms at the moment?&So do you have any ideas as to what might be the cause of your symptoms at the moment?\\
Have you noticed any changes in your weight?&Do you noticed any changes in your wit?\\
Okay have you noticed any mucus in your bowel motions?&Okay have you noticed any mucus in your bible Moshe?\\
\hline
\end{tabular}
\end{table*}

\begin{figure}[t]
\centering
\includegraphics[width=.65\textwidth]{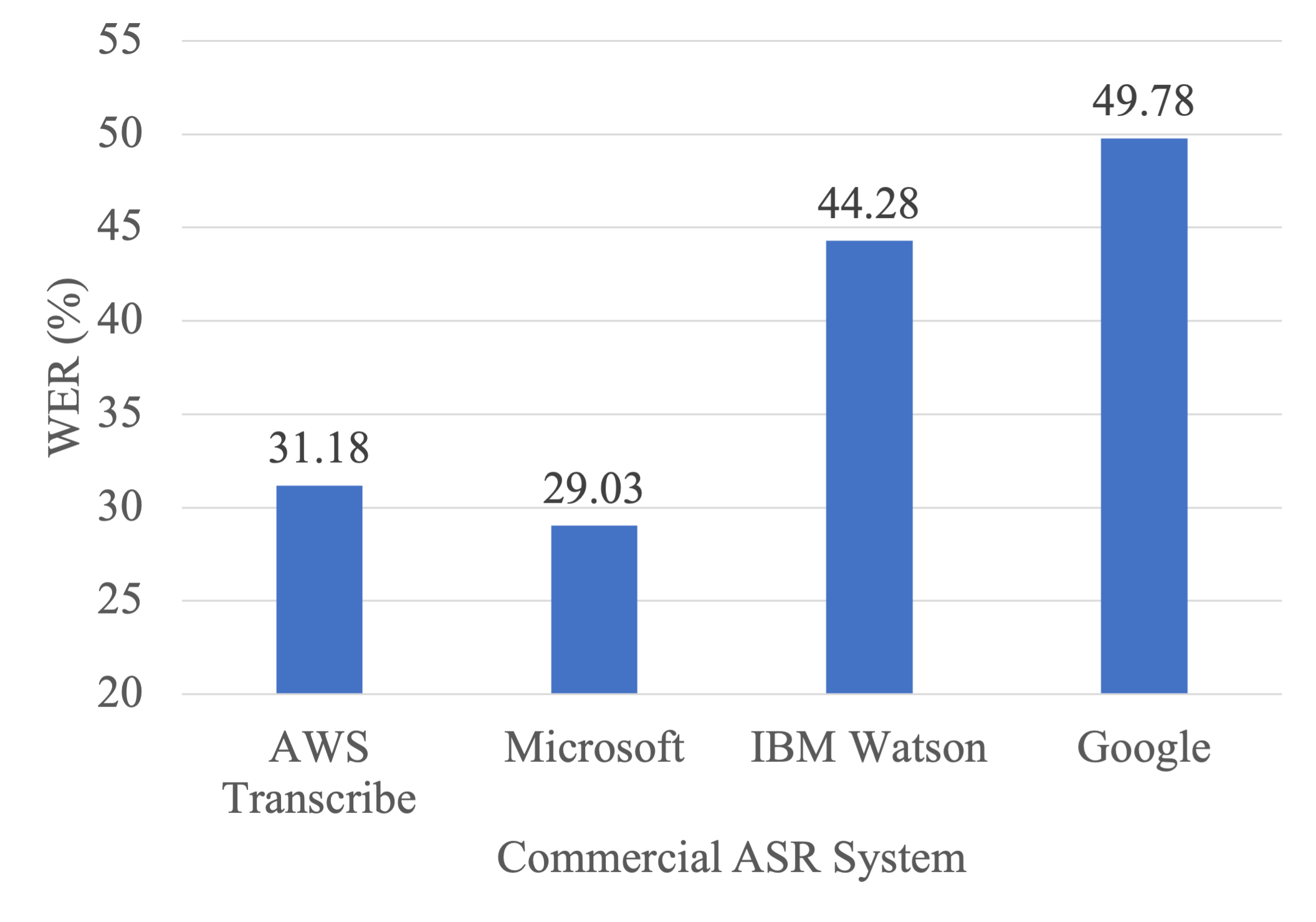}
\caption{Comparison of ASR Performance}
\label{fig:baselinewer}
\end{figure}

In Figure~\ref{fig:baselinewer} we plot the transcription error rate measured by Word Error Rate~(WER) for each ASR system. WER scores are calculated against the reference transcript of each audio file and the transcribed output from each ASR system. Accordingly, we find that Microsoft speech-to-text service generates the most accurate transcriptions from the GCD Dataset and Google speech-to-text service generates the least accurate. Although these are commercial ASR systems, the lack of knowledge on the medical domain terms and background noise may contribute to the performance differences seen in Figure~\ref{fig:baselinewer}. In the next section we present our post-error correction approach using seq2seq models. 

\section{Methods}
\label{sec:methods}

\begin{figure*}[t]
\centering
\includegraphics[width=0.8\textwidth]{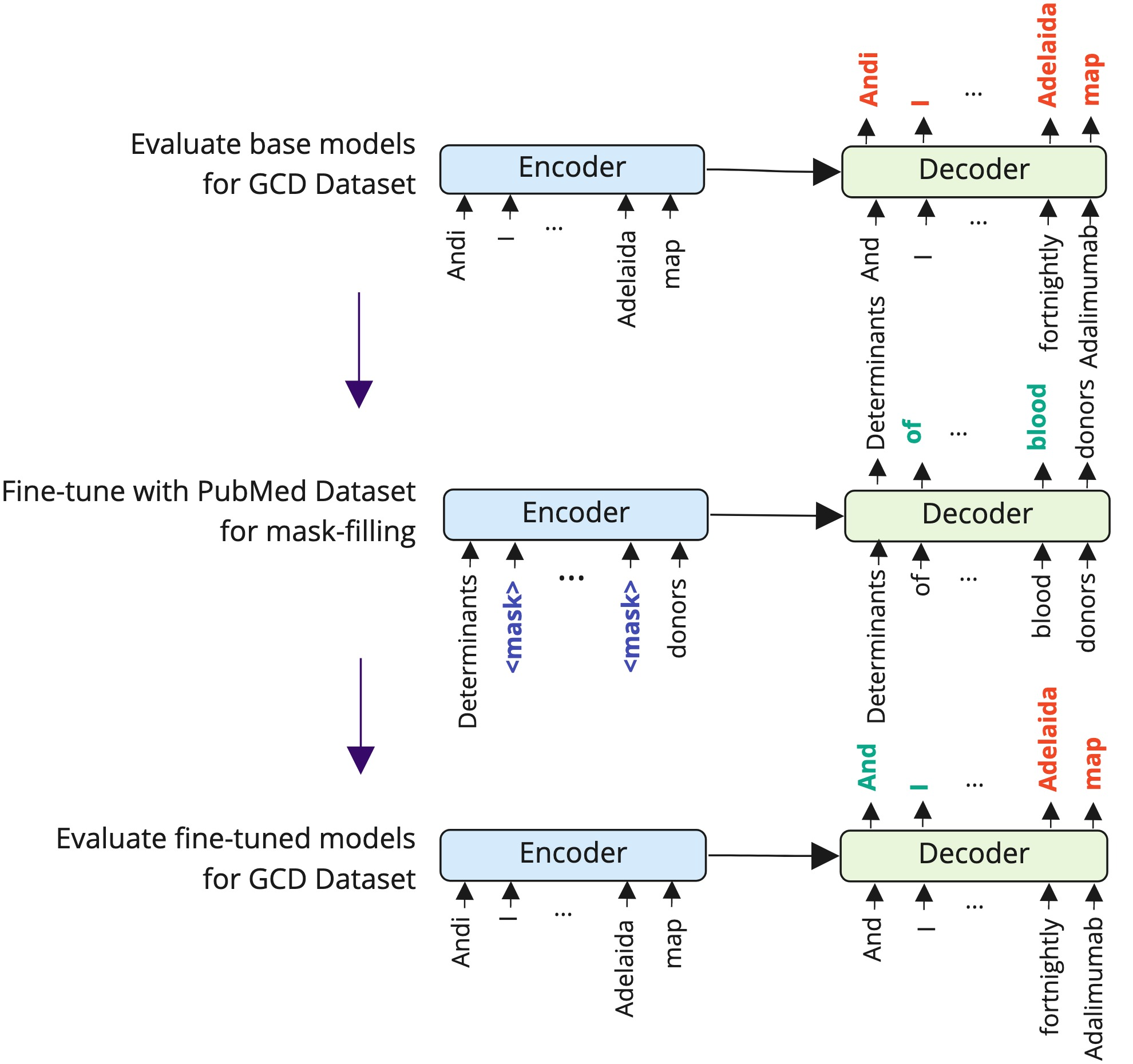}
\caption{Clinical Dialogue Transcription Error Correction using Seq2seq Model}
\label{fig:seqseq}
\end{figure*}

We view error correction as a seq2seq task performed using an Encoder-Decoder~(ED) architecture based language models. 
Here the text generated by the ASR forms the input to the encoder-part of the ED architecture, and the decoder-part is trained to generate the reference text. 
As illustrated in Figure~\ref{fig:seqseq}, a pre-trained language model needs also to be fine-tuned. The reasons for this are two fold: the ED models are general-purpose and not fine-tuned for terminology in the medical domain (i.e. vocabulary gap); and they are not fine-tuned to perform error-correction (i.e. objective gap). 
Essentially we need to integrate the domain vocabulary (e.g. gastrointestinal terminology) into the pre-trained language model that we want to use for error correction. 

As discussed in Section~\ref{sec:nhsdata} there is only a limited amount of data available for our clinical dialogue error correction task. 
Accordingly, it is challenging to use this dataset for both training (fine-tuning) and testing as we would in traditional machine learning. 
Instead we curate a dataset of Gastrointestinal text extracted from PubMed to fine-tune the general purpose language models on three distinct self-supervision tasks which are closer in nature to the error correction task. In this manner we aim to address the challenges posed by the gap in vocabulary and the difference in training objectives respectively.

\subsection{General Purpose Base Language Models}
\label{sec:basemodels}
Seq2seq learning has been performed using ED architectures created based on Recurrence~\cite{bahdanau2014neural,cho2014learning} and Transformers~\cite{lewis2019bart,T5}. Transformer based ED architectures are the state-of-the art and there exist different variants trained for different language modelling tasks. In this paper we consider the Bidirectional and Auto-Regressive Transformer~(BART~\cite{lewis2019bart}) and seq2seq Transfer Transformer~(T5~\cite{T5}) models, both of which have been pre-trained using large language corpora. 

\subsubsection{T5 Model}

T5 is a transformer-based architecture that is pre-trained on a mixture of seq2seq tasks~\cite{T5}. T5 architecture resembles the original transformer architecture~\cite{attention} with 12 transformer blocks in both the encoder and the decoder.
The T5 model is trained on the Colossal Clean Crawled Corpus~(C4) dataset, a collection of clean and natural English text from the internet in April 2019~(750GB in size). 
Multiple T5 model variants based on the number of attention blocks have been introduced~\cite{T5}. 
In this work, we fine-tune the T5-small, T5-base and T5-large models with 6, 12 and 24 attention blocks respectively. 

\subsubsection{BART}
The Bi-directional Auto-Regressive Transformer (BART) also uses the standard transformer architecture pre-trained for multiple seq2seq tasks. BART can be simplified as a transformer based architecture where the encoder is a generalised BERT~(i.e. bidirectional encoder)~\cite{devlin2019bert}, and the decoder is a GPT~(i.e. left-to-right decoder)~\cite{Radford2018ImprovingLU}. The BART model is trained on data from a combination of books and Wikipedia data, consisting of news, books, stories, and web text~(160GB in size). The model is trained by adding noise to text using an arbitrary noising function and learning the model to reconstruct the original text.

\subsection{PubMed Gastrointestinal Dataset}

PubMed is a collection of $\sim$33 million citations of biomedical literature collated from sources such as MEDLINE, life science journals and online books created by the US National Library for Medicine. It provides a search and retrieval engine for the public to extract biomedical articles. These articles either contains full-text and abstract text and are annotated by unique record identifiers called PMID. 
PubMed has been used as a resource for content classification~\cite{dernoncourt2017pubmed}, biomedical question and answering~\cite{jin2019pubmedqa} and biomedical entity recognition~\cite{lee2020biobert}. 

We also extract data from PubMed to create the PubMed Gastrointestinal Dataset. The main goal of extracting data from PubMed is to create a dataset introducing medical terminology to base language models. When choosing a dataset, there are differences between written and spoken English within specialist domains. The lack of availability of a larger spoken corpus has led us to use the written corpus, which is currently the most viable alternative. To this end, we adopted the following protocol to extract data from PubMed:
\begin{enumerate}
\item crawl PubMed to extract paper titles and abstracts. In this work we limit the search to articles related to gastrointestinal research. Accordingly the search terms are \textit{gastrointestinal symptoms,  diagnosis, clinical, examination, and patient};
\item clean titles by removing the Unicode characters; and
\item clean abstracts by removing the different Unicode characters, equations, figures, tables and URLs.
\end{enumerate}
After pre-processing we obtain a dataset with title and abstract pairs (see Table~\ref{tbl:pubmedsummary} for corpus statistics). The methods presented in this paper can be generalised to any medical domain by using the domain-specific search queries in the PubMed database search engine.
\begin{table}[t]
\centering
\renewcommand{\arraystretch}{1.1}
\caption{Summary of the PubMed Dataset}
\label{tbl:pubmedsummary} 
\begin{tabular}{ll}
\hline 
\textbf{Feature}&\textbf{Value}\\
\hline
Number of title, abstract pairs &\textbf{11,772}\\
Mean no. of words in title&\textbf{102}\\
Mean no. of words in abstract&\textbf{1,533}\\
\hline
\end{tabular}
\end{table}

\subsection{Fine-tune using Self-supervision}
\label{sec:finetunemodels}

The approach of using the same unsupervised data to create multiple training objectives is known as self-supervision. 
When fine-tuning the base language models we need to create self-supervisions tasks that have the general structure of an input-output text pair (i.e. seq-to-seq). 
When creating self-supervised datasets from PubMed data we are keen to empirically identify which fine-tuning task is best suited to introducing medical terminology as well as performing error-correction once fine-tuned. 
We created three variants of the PubMed dataset with a view to performing three different fine-tuning tasks on the base language models~\footnote{Accessible from the Hugging Face dataset repositories~\url{https://huggingface.co/gayanin}}. In Table~\ref{tbl:pubmedexamples} we present an example for each fine-tune task.

\begin{table*}[t]
\centering
\renewcommand{\arraystretch}{1.3}
\caption{Examples from the PubMed Gastrointestinal Dataset}
\label{tbl:pubmedexamples} 
\begin{tabular}{p{2cm}p{7cm}p{2.5cm}}
\hline 
\textbf{Task}&\textbf{Input}&\textbf{Output}\\
\hline
Summarisation&Helicobacter pylori is a worldwide infection. It is estimated that approximately 50\% of the general population is affected, but this percentage varies considerably between countries. \ldots\ This study confirms relatively high prevalence of H. pylori seropositivity among Italian healthy adults and points to sex, age, BMI and sociocultural class as persisting determinant features of H. pylori infection.&\multirow{6}{3cm}{Determinants of Helicobacter pylori seroprevalence among Italian blood donors.}\\

Paraphrasing&Determinants of seroprevalence of Helicobacter pylori among Italian blood donors&\\

Mask-filling&Determinants $<mask>$ $<mask>$ pylori $<mask>$ among Italian blood donors.&\\
\hline
\end{tabular}
\end{table*}

\begin{description}
\item [Summarisation] task generates a summary for a given text input. In the PubMed dataset, the abstract is considered as the input and the title is considered as the gold standard for the expected summary. No changes are required to the original PubMed dataset described in the previous section. Table~\ref{tbl:pubmedexamples} first row presents an example of abstract and title pair for summarisation from the PubMed dataset. 

\item [Paraphrasing] task generates a re-phrased text for a given text input. The goal of paraphrasing is to represent the meaning of a given text using different or re-arranged words. From the PubMed dataset, we used titles as the input to a T5 model fine-tuned for paraphrasing using the Google PAWS Dataset~\cite{zhang2019paws} to generate a paraphrased versions of the titles. The resulting dataset has title and paraphrased title pairs as shown in the example Table~\ref{tbl:pubmedexamples} second row. For fine-tuning using this dataset, we use the paraphrased title as the input and the title as the reference text.

\item [Mask-filling] is the task of predicting matching text for masked tokens in a text. To prepare the PubMed dataset for mask-filling we augment titles such that 25\% of the words in the title are replaced with the word $<mask>$. The resulting dataset has title and masked title pairs as shown in example in Table~\ref{tbl:pubmedexamples} third row. For the mask-filling fine tuning, we use the masked title as the input and the title as the reference text. 
\end{description}

\section{Evaluation}
\label{sec:eval}
The aim of this evaluation is two-fold. Firstly we measure the efficacy of base language models to perform error correction on clinical dialogue in Section~\ref{sec:baseeval}. 
Secondly, we identify which fine-tuning task is best for the clinical dialogue error correction task in Section~\ref{sec:finetuneeval}. 

\subsubsection{Performance Metric}
The metric selected to measure error correction is Word Error Rate~(WER). WER is derived from the Levenshtein distance which measures the differences between two strings. WER has been used as a performance metric in ASR systems~\cite{asr_review} and in post-ASR error correction~\cite{Mani2020TowardsUA,leng2021fastcorrect}. Given a language model output and a reference text, WER is calculated using Equation~\ref{eq:wer}. 
Here $S$, $D$ and $I$ refer to the number of substitutions, deletions, and insertions operations needed to transform the reference text to the language model output. 
$C$ refers to words that are equal in both reference and the output and $N$ refers to the number of words in the reference text and $N=S+D+C$. Lower WER scores are desirable. 

\begin{equation}
    WER = \frac{S+D+I}{N} = \frac{S+D+I}{S+D+C}
\label{eq:wer}
\end{equation}

\subsection{Comparison of Base Language Models}
\label{sec:baseeval}
We compare the base language models detailed in Section~\ref{sec:basemodels} for the task of clinical dialogue error correction. 
These models were pre-trained using large public domain language corpora for multiple language modelling tasks~(i.e. Wikipedia~\cite{lewis2019bart}, Book Corpus~\cite{lewis2019bart} and web extracted text~\cite{T5}). However they are not fine-tuned to the medical domain or error-correction task. We implement these models using the Python Hugging Face~\footnote{https://huggingface.co/} and PyTorch~\footnote{https://pytorch.org/} libraries while maintaining all default hyper-parameters. 
Models are evaluated using the four ASR outputs in the GCD datasets from Section~\ref{sec:nhsdata} on which mean WER is reported as a percentage.

\subsubsection{Results}
\begin{table*}[t]
\centering
\renewcommand{\arraystretch}{1.1}
\caption{Comparison of Base Language Models}
\label{tbl:basemodelresults} 
\begin{tabular}{llrrrr}
\hline 
\multicolumn{2}{c}{\textbf{Model}}&\multicolumn{4}{c}{\textbf{WER~(\%)}}\\
\textbf{Name} &\textbf{Version} & \textbf{AWS Transcribe} & \textbf{Microsoft} & \textbf{IBM Watson} & \textbf{Google}\\
\hline 
\multirow{3}{*}{T5}&T5-Small&\textbf{55.41}&\textbf{54.87}&\textbf{61.74}&\textbf{64.20}\\
&T5-Base&214.08&205.56&209.84&162.63\\
&T5-Large&163.96&163.54&153.64&137.19\\
\hline 
\multirow{2}{*}{BART}&BART-Base&\textbf{38.29}&\textbf{30.95}&\textbf{42.63}&\textbf{44.47}\\
&BART-Large&66.95&55.40&61.50&55.12\\
\hline
\end{tabular}
\end{table*}

Table~\ref{tbl:basemodelresults} represents the WER scores for T5 and BART model variants evaluated on each ASR output. Overall, smaller models~(i.e. less transformer blocks) achieve lower WER compared to larger models consistently across all four datasets. 
In T5 models, both \textit{base} and \textit{large} model variants have a WER score of more than 100\%. 
Models with a higher number of transformer blocks tend to generate longer sentences. 
Accordingly the number of \textit{insert} operations~(see Equation~\ref{eq:wer}) are higher when the output sentence is longer which results in a WER score higher than 100\%. Similar performance is observed with the two BART variants where larger model is producing higher WER due to output length. 
Between T5 and BART, the winner is BART-Base, although both fail to surpass the ASR WER scores~(see Figure~\ref{fig:baselinewer}). 
Accordingly, we will study the impact of fine-tuning just the T5-Small and BART-Base models. 

Table~\ref{tbl:output_examples} presents outputs generated for a sample input using T5 and BART base model variants. The sample input and its references text is randomly selected from the AWS Transcribe outputs and reference texts in the GCD dataset. The outputs from the T5 and BART models with increased number of transformer layers have evidently generated longer outputs which contributed to higher WER scores. Also, none of the models were able to correct the medical term present in the sample input.

\subsection{Comparison of Fine-tuned Language Models}
\label{sec:finetuneeval}
In this section, we compare the fine-tuned language models detailed in section~\ref{sec:finetunemodels} for the task of clinical dialogue error correction. These models are fine-tuned using the three self-supervising PubMed datasets we presented in Section ~\ref{sec:finetunemodels}. The hyper-parameters used in the fine-tuning were: optimiser is AdamW~\cite{loshchilov2017decoupled}; loss is cross-entropy; learning rate is $2\mathrm{e}-5$; and batch size is 16. 
For each fine-tuning task, the dataset was split 90/10 for training and evaluation. For each model the number of fine-tuning epochs varied between 10 and 40 as the fine-tuning was stopped when minimal evaluation loss was observed. For the summarisation task, the encoder input and decoder output sequence lengths were set to 1024 and 128 respectively; for paraphrasing and mask-filling tasks both encoder input and decoder sequence lengths were set to 512. Models are tested for error correction using the four ASR outputs on the GCD dataset from Section~\ref{sec:nhsdata} to report mean WER as a percentage.

\subsubsection{Results}

\begin{table*}[t]
\centering
\renewcommand{\arraystretch}{1.1}
\caption{Comparison of Fine-tuned Language Models}
\label{tbl:finemodelresults} 
\begin{tabular}{llrrrr}
\hline 
\multicolumn{2}{c}{\textbf{Model}}&\multicolumn{4}{c}{\textbf{WER~(\%)}}\\
\textbf{Name} &\textbf{Fine-tune Task} & \textbf{AWS Transcribe} & \textbf{Microsoft} & \textbf{IBM Watson} & \textbf{Google}\\
\hline 
\multirow{3}{*}{T5-Small}&Summarisation&63.39&66.89&69.44&73.80\\
&Paraphrasing&48.87&47.24&54.52&57.97\\
&Mask-filling&\textbf{38.83}&\textbf{35.86}&\textbf{45.16}&\textbf{46.87}\\
\hline 
\multirow{3}{*}{BART-base}&Summarisation&76.61&77.03&78.10&75.56\\
&Paraphrasing&43.31&37.46&47.51&49.48\\
&Mask-filling&\textbf{32.38}&\textbf{26.38}&\textbf{38.92}&\textbf{40.43}\\
\hline
\end{tabular}
\end{table*}

Table~\ref{tbl:finemodelresults} represents the WER scores for the fine-tuned model variants for the tasks summarisation, paraphrasing and mask-filling for each ASR system output.  Overall, mask-filling is the best performing fine-tuning task for both BART and T5 models across all four datasets. 
Summarisation task resulted in the highest WER scores, which makes it is an unsuitable fine-tuning task for error correction. High WER is caused by the difference between the input and output sequence lengths used in summarisation, whereby error correction expects similar lengths for both input and output. 
From paraphrasing and mask-filling tasks where the input and output sequences match, mask-filling has achieved the best performance. In addition to generating an output of the expected size, the model has learned to correct errors when fine-tuned for mask-filling. 
In fact, intuitively, mask-filling is most similar to error-correction of the three fine-tuning tasks where it is teaching the model to find missing words. 
Importantly, mask-filling has improved the BART models to outperform ASR performance~(see Figure~\ref{fig:baselinewer}) with the Microsoft, IBM Watson and Google datasets and to perform comparably with the AWS Transcribe dataset. 

\begin{table*}[t]
\centering
\renewcommand{\arraystretch}{1.3}
\caption{Examples from Language Model Variants}
\label{tbl:output_examples} 
\begin{tabular}{p{3cm}p{9cm}}
\hline
\rowcolor{gainsboro}
&\textbf{Output}\\
\hline
Reference Text & \textit{And I know that youve been on fortnightly Adalimumab}\\
Input & Andi I know that youve been on fortnightly Adelaida map?\\
\hline
\rowcolor{gainsboro}
\textbf{Baseline Models}&\\
\hline
T5-Small & Andi I know that youve been on the fortnightly Adelaida map \\
T5-Base& been on fortnightly Adelaida map. Andi I know that youve been tagged as andi Im sure thats where youre on, but im not sure if its just me or am I right?\\
T5-Large& youve been on the fortnightly Adelaida map been on fortnightly Adelaida map. Andi I know that youve been tagged as andi Im sure thats where youre on, but im not sure if its just me or am I right?\\
BART-Base & Andi I know that youve been on fortnightly Adelaida maps\\
BART-Large &AndAll of the time.Andi I know that youve been on fortnightly \\
\hline
\rowcolor{gainsboro}
\textbf{Fine-Tuned Models} &\\
\hline
T5-Summarisation &I know that youve been on fortnightly Adelaida map \\
T5-Paraphrase &Andi I know that youve been on the \textit{two-weekly} Adelaida map. \\
T5-Mask-filling &Andi I know that youve been on fortnightly Adelaida map. \\
BART-Summarisation & Adelaida map on fortnightly basis. \\
BART-Paraphrase & I know youve been on the fortnightly Adelaida map.\\
BART-Mask-filling & \textit{And I know} that youve been on fortnightly Adelaida map\\
\hline
\end{tabular}
\end{table*}

Table~\ref{tbl:output_examples} presents outputs generated for a sample input using fine-tuned models. The models fine-tuned for summarisation generate shorter text which resulted in increased WER scores. In comparison, models fine-tuned for paraphrasing and mask-filling are generating text that are comparable to the input in length. Moreover, we observe several improvements to the output text such as T5-paraphrasing replacing the word ``fortnightly'' with ``two-weekly'' and BART Mask-filling accurately improving ``Andi I know'' to ``And I know''. However, even after fine-tuning for medical terminology, the models struggle to find the phonetic similarity between the ASR output and medical terminology. For example models fail to correct ``Adelaida map'' to  ``Adalimumab'' which will be a focus area for us to improve in future work. 

\section{Discussion}
\label{sec:discuss}

\begin{table*}[t]
\centering
\renewcommand{\arraystretch}{1.1}
\caption{Comparison of ASR and Language Model Outputs for Error Correction}
\label{tbl:discuss} 
\begin{tabular}{llrrrr}
\hline 
\multirow{2}{*}{\textbf{Model}}&\multirow{2}{*}{\textbf{Data}}&\multicolumn{4}{c}{\textbf{Number of Instances}}\\
&& \textbf{AWS Transcribe} & \textbf{Microsoft} & \textbf{IBM Watson} & \textbf{Google}\\
\hline 
\multirow{2}{*}{ASR}&Equal&30&27&7&1\\
&Different&269&273&277&224\\
\hline
\multirow{2}{*}{T5 for Mask-filling}&Equal&7&6&2&1\\
&Different&292&294&282&224\\
\hline
\multirow{2}{*}{BART for Mask-filling}&Equal&1&1&0&1\\
&Different&298&299&284&224\\
\hline
\end{tabular}
\end{table*}

The aim of this exploratory evaluation is to find out the type of utterances that contributed to the performance improvements with BART fine-tuned for mask-filling. To this end, first we split each ASR output in the GCD dataset in to data instances that ASR correctly transcribed~(ASR output equal to reference text) and incorrectly transcribed. Then we evaluate each subset of data using the best performing models we found in Section~\ref{sec:finetuneeval} which are T5 and BART models fine-tuned for mask-filling. In Table~\ref{tbl:discuss} we present the results in which we opt to present the number of utterances instead of WER for easy interpretation. 

In the ASR row we present the baselines where \textit{Equal} row refers to the number of utterances correctly transcribed by ASR and \textit{Different} row refers to the number of utterances incorrectly transcribed by ASR. For example, Google speech-to-text ASR system only transcribed one utterance correctly and 224 utterances had some differences when compared to the reference text. There is a difference between the total number of audio utterances of 329~(from Table~\ref{sec:nhsdata}) and Google ASR outputs. For given audio file, ASR system is generating a different number of utterances compared to the reference text. This is because, the ASR system is skipping some audio content that it cannot transcribe with a level of confidence. 
Evidently, ASR systems with higher error have less number of total utterances in Table~\ref{tbl:discuss}. 

At first glance, we find that the number of equal utterances are reduced with T5 and BART models and the number of different utterances are increased. But a close examination reveals that the performance improvements we observed in Section~\ref{sec:finetuneeval} over ASR system is due to making less errors within utterances that are different. This is clearly seen when comparing the number of utterances for Google where the numbers are not different from ASR yet achieved a performance improvement of 9.35\%~(49.78 - 40.43) and 2.91\%~(49.78 - 46.87) with the BART and T5 models. 
However we find that our models are introducing errors to utterances that ASR systems have correctly transcribed. A methodology to mitigate this will be explored in future work. 

\section{Conclusion}
\label{sec:conc}

In this paper, we presented a seq2seq learning approach for clinical dialogue transcription error correction. 
Given the lack of clinical dialogue data, we presented an approach which uses public medical domain data to fine-tune a language model to introduce domain specific clinical terms. 
We found out that PubMed data from a specific domain can be used in self-supervised manner to create data to fine-tune a general purpose seq2seq model.
Importantly our results suggest that the choice of fine-tuning task has a significant impact on the post-ASR error correction task. 
Specifically we found that the mask-filling was closely aligned to the target transcription error correction task compared to alternative fine-tuning tasks such as summarisation or paraphrasing.
With this method, we were able to surpass the performance of three out of four commercial ASR systems on a comparative study with fine-tuned T5 and BART seq2seq models.
In future work, we plan to introduce new fine-tune tasks with more self-supervised data to improve model knowledge of phonetic relationships between medical and regular phrases and improve error correction performance.

\section*{Acknowledgements}
We would like to thank the National Health Service (NHS) Grampian, including Amy Kousha, John Thomson and Jill Ferbrache, who helped to curate the IBD clinic role-playing dialogues.

\bibliographystyle{splncs04}
\bibliography{references}

\end{document}